\documentclass[letterpaper, 10 pt, conference]{ieeeconf}  % Comment this line out

\usepackage[dvips]{graphicx}
\usepackage{amsmath}
\usepackage{amsfonts}
\usepackage{amssymb}
\IEEEoverridecommandlockouts                              % This command is only
\title{\LARGE \bf
Feasibility of random basis function approximators\\ for modeling
and control}
\author{Ivan Yu. Tyukin and Danil V. Prokhorov% <-this % stops a space
\thanks{Ivan Tyukin is with Department of Mathematics,
        University of Leicester, UK, Department of Automation and
        Control Processes, St. Petersburg State University of
        Electrical Engineering, Russia, Perceptual Dynamics Lab,
        RIKEN BSI, Japan
        {\tt\small I.Tyukin@leicester.ac.uk}}%
\thanks{Danil Prokhorov is with Toyota Research Institure NA, Ann Arbor, MI 48105, USA
        {\tt\small dvprokhorov@gmail.com}}%
}

\newcommand{\Real}{\mathbb{R}}

\begin{document}

\maketitle
\thispagestyle{empty}
\pagestyle{empty}

%%%%%%%%%%%%%%%%%%%%%%%%%%%%%%%%%%%%%%%%%%%%%%%%%%%%%%%%%%%%%%%%%%%%%%%%%%%%%%%%
\begin{abstract}

We discuss the role of random basis function approximators in
modeling and control.  We analyze the published work on random basis
function approximators and demonstrate that their favorable error
rate of convergence $O(1/n)$ is guaranteed only with very
substantial computational resources.  We also discuss implications
of our analysis for applications of neural networks in modeling and
control.

\end{abstract}

%%%%%%%%%%%%%%%%%%%%%%%%%%%%%%%%%%%%%%%%%%%%%%%%%%%%%%%%%%%%%%%%%%%%%%%%%%%%%%%%
\section{INTRODUCTION}

Efficient modeling and control of complex systems in the presence of
uncertainties is important for modern engineering. This is
especially true in the domain of intelligent systems that are
designed to operate in uncertain environments. Uncertainties in such
systems are usually quantitative relations (maps) between measured
signals
\[
\begin{split}
& x_1(t),x_2(t),\dots,x_d(t) \mapsto f(x_1(t),x_2(t),\dots,x_d(t)),\\
& \ \ \ \ \ \ \ \ \ x_i:\Real\rightarrow\Real,
\end{split}
\]
and the number of these signals may be large.

Physical models of such relations $f(\cdot)$ are not always
available, and it is quite common to use mathematical substitutes
such as, e.g., superpositions of (basis) functions that are capable
of approximating a-priori unknown  $f(\cdot)$ with the required
precision. Thus, successful modeling and control in the domain of
intelligent systems are critically dependent on availability of
adequate and efficient function approximators which can take care of
various uncertainties in the system.

In the domain of modeling and control of intelligent systems the
multilayer perceptrons (MLP) and radial basis functions (RBF)
networks are popular function approximators \cite{Haykin99}. The MLP
uses a basis in the form of the sigmoids with {\em global} support.
For one-hidden layer MLP, its output is determined by
\begin{equation}\label{eq:mlp}
f_n(x)=\sum_{i=1}^n c_i \frac{1}{1+e^{(w_i^{T}x+b_i)}}
\end{equation}
Typically, both nonlinear ($w_i$ and $b_i$) and linear ($c_i$)
parameters, or weights, are subject to training on data specific to
the problem at hand (full network training).

The RBF networks use a basis in the form of the Gaussians with {\em
local} (but not compact) support:
\begin{equation}\label{eq:rbf}
f_n(x)=\sum_{i=1}^n c_i e^{(-\|w_i^{T}x+b_i\|^2)}
\end{equation}
Though all parameters may be trained in principle, typically only
linear weights $c_i$ of the RBF network are trained. The locations
and the widths of the Gaussians are usually set on a uniform or
nonuniform grid covering the operating domain of the system.

The popularity of approximators (\ref{eq:mlp}), (\ref{eq:rbf}) is
not only due to their approximation capabilities (see e.g.,
\cite{Cybenko}, \cite{Hornik90}, \cite{Park93}) and their homogenous
structure but also due to efficiency of approximators
(\ref{eq:mlp}), (\ref{eq:rbf}) in high dimensions. In particular, if
{\it all parameters $w_i$, $c_i$, $b_i$ } are allowed to vary, the
rate of convergence of the approximation error of a target function
$f\in\mathcal{C}^{0}[0,1]^d$ as a function of $n$
(the number of elements in the network) is shown to be independent
of the input dimension $d$ \cite{Jones:1992}, \cite{Barron}.
% where is it shown that the target function must be increasingly smooth with the increase
%of input dimensions for the independence to hold?
Furthermore, the achievable rate of convergence of the $L_2$-norm of
$f(x)-f_n(x)$ is shown to be of order $O(1/n)$.

Despite these advantageous features of approximators (\ref{eq:mlp}),
(\ref{eq:rbf}), viz. favorable independence of the convergence rates
on the input dimension of the function to be approximated, the issue
is how to achieve the convergence rate of order $O(1/n)$ in
practice. Even though \cite{Jones:1992}, \cite{Barron} offer a
constructive procedure for optimal selection of basis functions,
each step of these procedures involves a nonlinear optimization
routine searching for the best possible values of $w_i$, $b_i$ (see
Section II for details). It is also shown in \cite{Barron} that if
only linear parameters of (\ref{eq:mlp}), (\ref{eq:rbf}) are
adjusted the approximation error {\it cannot be made} smaller than
$1/n^{2/d}$ uniformly for functions satisfying the same smoothness
constraints.

The necessity to adjust nonlinear parameters of (\ref{eq:mlp}),
(\ref{eq:rbf}), restricts practical application of these models to
those problems in which such optimization is feasible.  Adaptive
control with nonlinearly parameterized models remains a challenging
issue; see e.g., \cite{tpt2002_at}, \cite{tpt2003_tac},
\cite{tpvl2007_tac}, \cite{Annaswamy99}.

%THIS PARAG. ADD NOTHING: NEXT TWO PARAG. DESCRIBE IT WELL!  Because
%of these theoretical limitations concerning nonlinearly
%parameterized models in the domains of modeling, control and
%identification the following question arises: it it is possible to
%avoid adjustment of nonlinear parameters of (\ref{eq:mlp}),
%(\ref{eq:rbf}) while maintaining the convergence rates of an
%approximation scheme {\it close in some sencse (but not precisely
%same)} to $O(1/n)$ and ensuring {\it practical} independence on $d$?
%A framework to address this issue was proposed in \cite{igelpao95}.

Though published quite a while ago, the paper by Igelnik and Pao
\cite{igelpao95} (see also comments \cite{lichow97}) has recently received numerous citations in a
variety of intelligent control publications;  see, e.g.,
\cite{he05}--\cite{Liuetal07}. The paper advocates the use of
random basis in the MLP (\ref{eq:mlp}) and RBF (\ref{eq:rbf})
networks. That is, the nonlinear parameters $w_i$ and $b_i$ are to
be set randomly at the initialization, rather than through training.
The only trainable parameters are those which enter the network
equation linearly ($c_i$).

The paper \cite{igelpao95} provides mathematical justification to
the use of linear-in-parameters function approximators for modeling
and control, crucially simplifying analysis of properties of the
closed-loop control system featuring such approximators.  While
analysis simplification is attractive, it entails a number of issues
which are important to consider whenever planning to apply such
random basis function approximators in practice. We show that the
rate of convergence of order $O(1/n)$ for such approximators is
achievable only for large $n$, and it is probabilistic in nature.
The latter feature may require introduction of a supervisory
mechanism in the control system to re-initialize the network if the
required accuracy is not met.

The paper is organized as follows. In Section \ref{sec:Analysis} we
analyze the reasoning in \cite{igelpao95} and  compare these results
with  \cite{Jones:1992}, \cite{Barron}. We show that, although these
results may seem inconsistent (i.e., the lower approximation bound
$1/n^{2/d}$ derived in \cite{Barron} for any linear-in-parameter
approximator vs. the rate of convergence of order $O(1/n)$ and
independent of $d$ in \cite{igelpao95}), they are derived for
different asymptotics (for every $n$ in \cite{Jones:1992},
\cite{Barron} vs. for large $n$ in \cite{igelpao95}) and use
different convergence criteria (deterministic in \cite{Jones:1992},
\cite{Barron} vs. statistical in \cite{igelpao95}).  Implications of
our analysis are illustrated in Section \ref{sec:Example}  with a
simple example, followed by a discussion in Section
\ref{sec:Discussion}.  Section \ref{sec:Conclusion} concludes the
paper.

%%%%%%%%%%%%%%%%%%%%%%%%%%%%%%%%%%%%%%%%%%%%%%%%%%%%%%%%%%%%%%%%%%%%%%%%%%%%%%%%
\section{Function Approximation Concepts}\label{sec:Analysis}

In this section we review and compare two results for function
approximation with neural networks.  The first result is the
so-called {\it greedy approximation} upon which the famous Barron's
construction is based \cite{Barron}. In this framework a function is
approximated by a sequence of linear combinations of basis
functions. Each basis function is to satisfy certain optimality
condition, and as a result the overall rate of convergence is
optimized as well.

The second result is the random basis function approximator also
known as the Random Vector Functional-Link (RVFL) network
\cite{igelpao95} in which the basis functions are randomly chosen,
and only their linear parameters are optimized.
% not clear what the final approximation means
Both results enjoy the convergence rates that do not depend on the
input dimension $d$ of the target functions. However, there are
differences important for practical use of these results.

First, as we show below the number of practically required
approximation elements (network size) that guarantees given
% in the expansions of what?
approximation quality differ substantially. Second, the quality
criteria are also different: in the framework of greedy
approximation this is merely the $L_2$-norm which is a deterministic
functional, whereas in the RVFL framework the criterion is {\it
statistical}.

\subsection{Approximation problem}

Consider the following class of problems. Let $f:[0,1]^d\subset
\Real^d\rightarrow\Real$ be a continuous function, and
\[
\|f\|^2=\langle f,f \rangle=\int_{[0,1]^d} f(x)f(x) d x,
\]
be the $L_2$-norm of $f$.  Suppose that
$g:\Real\rightarrow\Real$ be a function such that
\[
\|g(\cdot)\|\leq M, \ M\in\Real_{>0},
\]
and that
\[
f\in \overline{\mathrm{convex} \ \mathrm{hull}} \ \{g(w^{T}x + b)\}, \ w\in\Real^d, \ b\in\Real,
\]
In other words there is a sequence of $w_i$, $b_i$, and $c_i$ such that
\[
f(x)=\sum_{i=1}^{\infty} c_i g(w_i^{T}x + b_i), \ \sum_{i=1}^{\infty}c_i=1
\]
Let
\begin{equation}\label{eq:approximation}
f_n(x)=\sum_{i=1}^n c_i g(w_i^{T}x+b_i)
\end{equation}
be a superposition of functions $g(w_i^{T}x+b_i)$. The question is
how many elements do we need to pick in (\ref{eq:approximation}) to
assure that the approximation error does not exceed certain
specified value?

\subsection{Greedy approximation and Jones Lemma}

In order to answer the question above one needs first to determine
the error of approximation. It is natural for functions from $L_2$
to define the approximation error as follows:
\begin{equation}\label{eq:error_greedy}
e_n=\|f_n - f\|
\end{equation}

The classical Jones iteration \cite{Jones:1992} (refined later by
Barron \cite{Barron}) provides us with the following estimate of
achievable convergence rate:
\begin{equation}\label{eq:rate:greedy}
\begin{split}
e_n^2&\leq \frac{M'^2 e_0^2}{n e_0^2 + M'^2}, \  M'\in\Real_{>0},\\
M'&> \sup_{g} \|g\|+\|f\|.
\end{split}
\end{equation}
The rate of convergence depends on $d$ only through the $L_2$-norms
of $f_0$, $g$, and $f$. The iteration itself is deterministic and
can be described as follows:
\begin{equation}\label{eq:Jones_iteration}
\begin{split}
f_{n+1}&=(1-\alpha_n)f_n + \alpha_n g_n\\
\alpha_n&= \frac{e_n^2}{M''^2 + e_n^2}, \ \ M''>M'
\end{split}
\end{equation}
where $g_n$ is chosen such that the following condition holds
\begin{equation}\label{eq:Jones_iteration:g_n}
\langle f_n-f,g_n-f\rangle < \frac{((M'')^2-(M')^2) e_n^2}{2(M'')^2},
\end{equation}
This choice is always possible (see \cite{Jones:1992} for
details).

According to (\ref{eq:rate:greedy}) the rate of convergence of  such
approximators is estimated as
\[
\|e_n\|^2 =  O(1/n).
\]
% why =, not \sim?
This convergence estimate is {\it guaranteed} because it is the
upper bound for the approximation error at the $n$th step of
iteration (\ref{eq:Jones_iteration}).

%The first one is the result by Barron \cite{barron} on achievable rates of convergence of the
%function approximation schemes with adjustable basis in $L_2$. Specifically, we refer to the
%fact that the $L_2$-norm of the approximation error can be made bounded from above as follows:

\subsection{Approximation with randomly chosen basis functions}

We now turn our attention to the result in \cite{igelpao95}. In this
approximator the original function $f(\cdot)$ is assumed to have the
following integral representation\footnote{We keep the original
notation of \cite{igelpao95} which uses both $\omega$ and $w$ for
the sake of consistency.}
\begin{equation}\label{eq:approximation:2}
f(x)=\lim_{\alpha\rightarrow\infty}\lim_{\Omega\rightarrow\infty} \int_{W^d} F_{\alpha,\Omega}(\omega)g(\alpha w^{T}x+b)d \omega,
\end{equation}
%  is W^d volume in x or number of approx. elements?
where $g:\Real\rightarrow\Real$ is a non-trivial function from $L_2$:
\[
0<\int_{\Real}g^2(s)ds < \infty,
\]
where
$\omega=(y,w,u)\in\Real^{d}\times\Real^d\times[-\Omega,\Omega]$, $\Omega\in\Real_{>0}$, $W^d=[-2d\Omega;2d\Omega]\times I^d \times V^d$, $V^d=[0;\Omega]\times[-\Omega;\Omega]^{d-1}$, $b=-(\alpha w^{T} y + u)$ and
\[
F_{\alpha,\Omega}(\omega)\sim \frac{\alpha\prod_{i=1}^d w_i}{\Omega^d 2^{d-1}} f(y).
\]
See \cite{igelpao95}, \cite{lichow97} for more detailed description.
Function $g(\cdot)$ induces a parameterized basis. Indeed if we were
to take integral (\ref{eq:approximation:2}) in quadratures for
sufficiently large values of $\alpha$ and $\Omega$, we would then
express $f(x)$ by the following sums of parameterized $g(\alpha
w^{T}x+b)$ \cite{igelpao95}:
\begin{equation}\label{eq:approximation:R}
%f_n(x)\approx \sum_{i,j,m}^n c_{i} g(\alpha w_i^{T}x + b_i), \ b_i=- \alpha(w_i^{T} y_i + u_i)
f_n(x)\approx \sum c_{i} g(\alpha w_i^{T}x + b_i), \ b_i=-
\alpha(w_i^{T} y_i + u_i)
\end{equation}
The summation in (\ref{eq:approximation:R}) is taken over points
$\omega_i$ in $W^d$, and $c_i$ are weighting coefficients.
% NO NEED TO REDEFINE \omega.
Variables $\alpha$ in (\ref{eq:approximation:2}) and $\alpha_n$ in
(\ref{eq:Jones_iteration}) play different roles in each
approximations schemes. In (\ref{eq:Jones_iteration}) the value of
$\alpha_n$ is set to ensure that the approximation error is
decreasing with every iteration, and  in (\ref{eq:approximation:2})
it stands for a scaling factor of random sampling.

%{\bf Ivan, I understand that some confusion of notation (as
%indicated by reviewers) comes from the original paper by Igelnik and
%Pao.  But could you still explain their notation?  What is
%$\omega=(y,w,u)$ and $\alpha$?  Can we remove $\alpha$ altogether,
%or is it related to $\alpha_n$ in the previous section?}

The main idea of \cite{igelpao95} is to approximate integral
representation (\ref{eq:approximation:2}) of $f(x)$ using the Monte-Carlo integration method as
\begin{equation}\label{eq:MC_approximation}
\begin{split}
f(x)&\sim \frac{4 \Omega^d}{n} \lim_{\alpha\rightarrow\infty}\lim_{\Omega\rightarrow\infty} \sum_{k=1}^n F_{\alpha,\Omega}(\omega_k)g(\alpha w^{T}_k x+b_k)\\
&=\frac{4}{n} \lim_{\alpha\rightarrow\infty}\lim_{\Omega\rightarrow\infty} \sum_{k=1}^n c_{k,\Omega}(\alpha,\omega_k)g(\alpha w^{T}_kx+b_k)\\
&=f_{n,\omega,\Omega}(x),
\end{split}
\end{equation}
where the coefficients $c_k(\alpha,w_k)$ are defined as
\begin{equation}\label{eq:MC_coefficients}
c_{k,\Omega}(\alpha,w_k)\sim \frac{\alpha\prod_{i=1}^d w_{k,i}}{2^{d-1}} f(y_k)
\end{equation}
and $\omega_k=(y_k,w_k,u_k)$ are randomly sampled in $W^d$ (domain
of parameters, i.e., weights and biases of the network).

When the number of samples, $n$, i.e., {\em the network size}, is
% I emphasized the network size to avoid confusion with more common notion of the training sample size
large, then the expectation $E_\omega(n,x)$
\[
\begin{split}
E_\omega(n,x)&=f(x)-\frac{4}{n}\sum_{k=1}^n c_{k,\Omega}(\alpha,\omega_k)g(\alpha w^{T}_kx+b_k)
\end{split}
\]
converges to zero for large $n$ (Theorem 1 in \cite{igelpao95}):
\[
\lim_{n\rightarrow\infty} E_\omega(n,x) =0.
\]

The advantage of the Monte-Carlo integration, and hence the
approximation techniques that are based upon this method, is its
order of convergence for large $n$. It is known that if $W^d$ is
bounded (i.e., its volume is bounded) then {\it the variance} of the
estimate (\ref{eq:MC_approximation}) is bounded pointwise from
above:
\begin{equation}\label{eq:rate:MC}
\begin{split}
\mathrm{Var}_\omega
(n,x)&=\lim_{n\rightarrow\infty}|W^d|\frac{\sigma^2_f(x)}{n}
\end{split}
\end{equation}
where
\[
\sigma^2_f(x)=\int_{W^d} (c_{k,\Omega}(\alpha,\omega)g(\alpha w^{T}_kx+b_k)-f(x))^2d\omega.
\]
In this sense the order of Monte-Carlo approximation for large
number of processing elements of the approximator (network nodes)
$n$ may be made similar to that of the greedy approximation.

\subsection{Comparison}

There are, however, important points that make this method different
from the greedy approximation:
\begin{itemize}
\item  Approximation ``error'' (\ref{eq:rate:MC}) is statistical, whereas the approximation
error (\ref{eq:error_greedy}) is deterministic. This means that
     $f_{n,\omega,\Omega}(x)$ is not at all guaranteed  to be close to $f(x)$
     for {\it every} randomly chosen set of length $n$.
% sample replaced by set to avoid confusion with training sample (reviewers might be confused if they read too quickly)
     We can, however, conclude that for sufficiently small $\gamma=\sigma^2_f(x)/(N \varepsilon)^2$
     the probability that $f_{n,\omega,\Omega}(x)$ is close to $f(x)$ approaches $1$ (from the Chebyshev inequality):
     \[
     \begin{split}
     &P\left(\left|\frac{4}{n}\sum_{k=1}^n c_{k,\Omega}(\alpha,\omega_k)g(\alpha w^{T}_kx+b_k)-f(x) \right|<\varepsilon\right)\\
     & \ \ \ \ \ \ \ \ \geq 1-\gamma
     \end{split}
     \]
% why P >= 1-\gamma, why not 1 >= P >= 1-\gamma?

\item For the Monte-Carlo based scheme (\ref{eq:MC_approximation})--(\ref{eq:rate:MC}) to converge, one needs
to ensure that $W^d$ is bounded. This, however, conflicts with the
requirement that $\Omega\rightarrow\infty$
(\ref{eq:approximation:2}). Hence the class of functions to which
the scheme applies is restricted. In order to mitigate this
restriction, it is proposed to consider functions $g(\cdot)$ with
compact support, and for this class of functions
dimension-independent (statistical) rate of convergence
(\ref{eq:rate:MC}) is guaranteed.
\item The relatively fast rate of convergence (\ref{eq:rate:MC}) is guaranteed only for large $n$.
% THIS STATES THE VERY SAME FACT AS ABOVE (DUPLICATE SENTENCE):
%In practice, this means that one needs to pick relatively large $n$
%to ensure that the estimate (\ref{eq:rate:MC}) holds.
\end{itemize}

These points are summarized in Table 1.

\begin{table}
\caption{Greedy vs Random Vector Functional-Link approximators}

  \begin{tabular*}{\columnwidth}{@{\extracolsep{\fill}}|c|c|c|}
    \hline
    &  & \\
     Feature & Greedy Approximation & Random \\
    & &  \\
    \hline
      &  & \\
   Quality criterion &  Deterministic: & Statistical:\\
    &  & \\
     & $e_n=\|f_n-f\|$ & $ e_n=\sqrt{\mathrm{Var}_{\omega}(n,x)}$ \\
     &  & \\
    \hline
   &   & \\
Convergence rate   & $e_n^2\leq O(1/n)$  & $e_n^2\leq O(1/n)$  \\
 &  & \\
   & for all $n\geq 1$ & for large $n$\\
   &  & \\
 \hline
  \end{tabular*}
\end{table}

\section{Example}\label{sec:Example}

\begin{figure}[!h]
\begin{center}
\includegraphics[width=0.85\columnwidth]{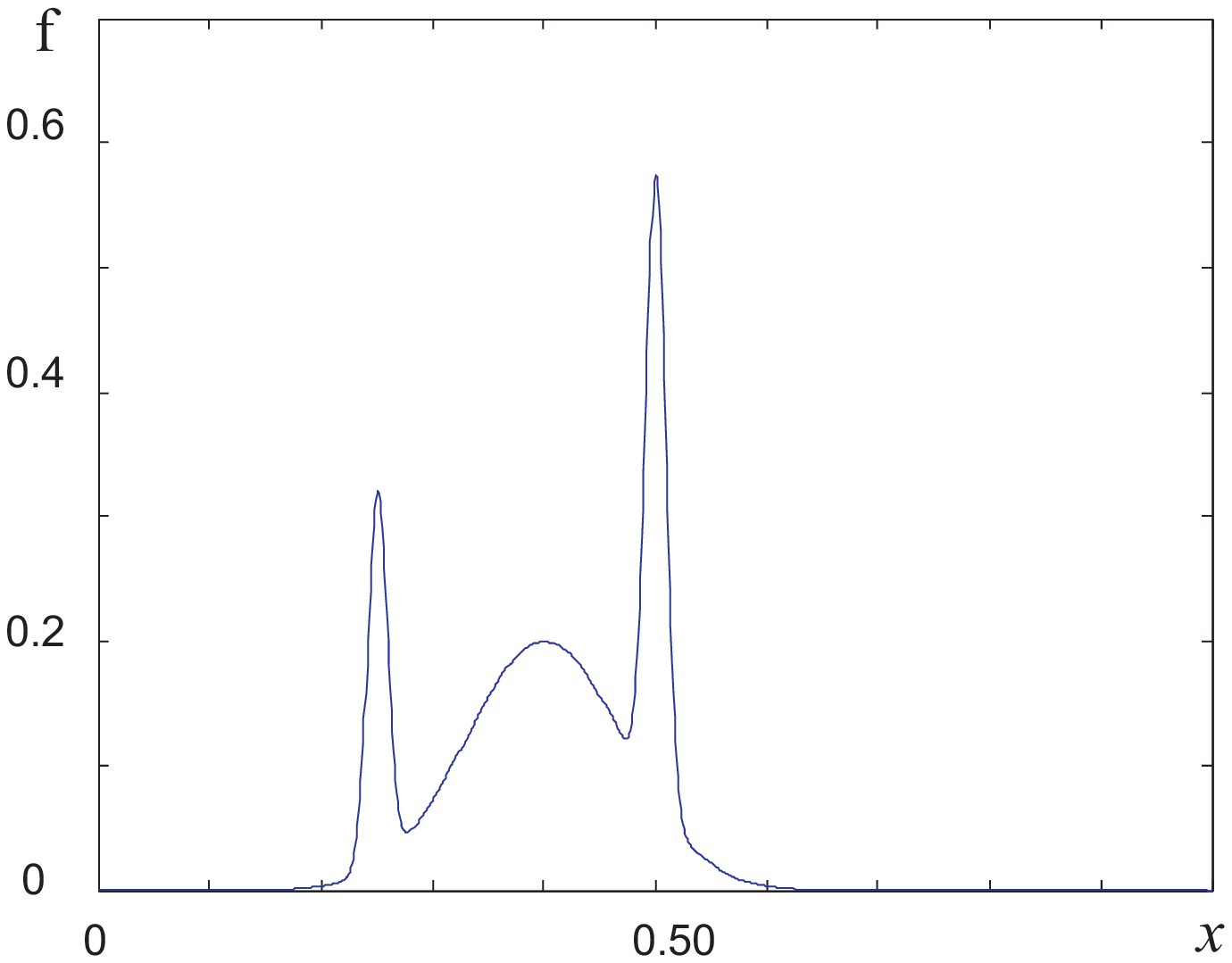}
\vskip 6mm
\includegraphics[width=0.85\columnwidth]{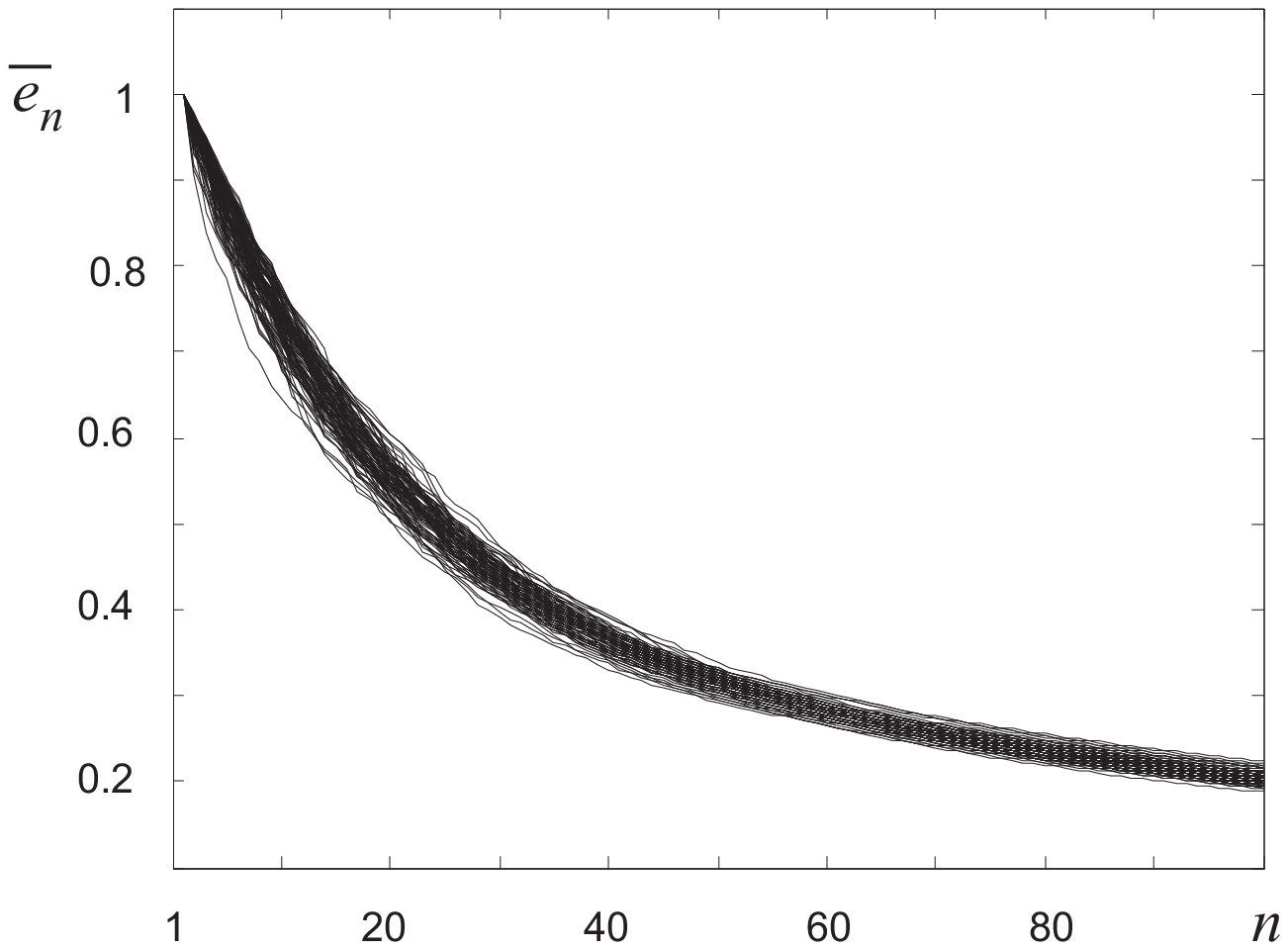}
\vskip 6mm
\includegraphics[width=0.85\columnwidth]{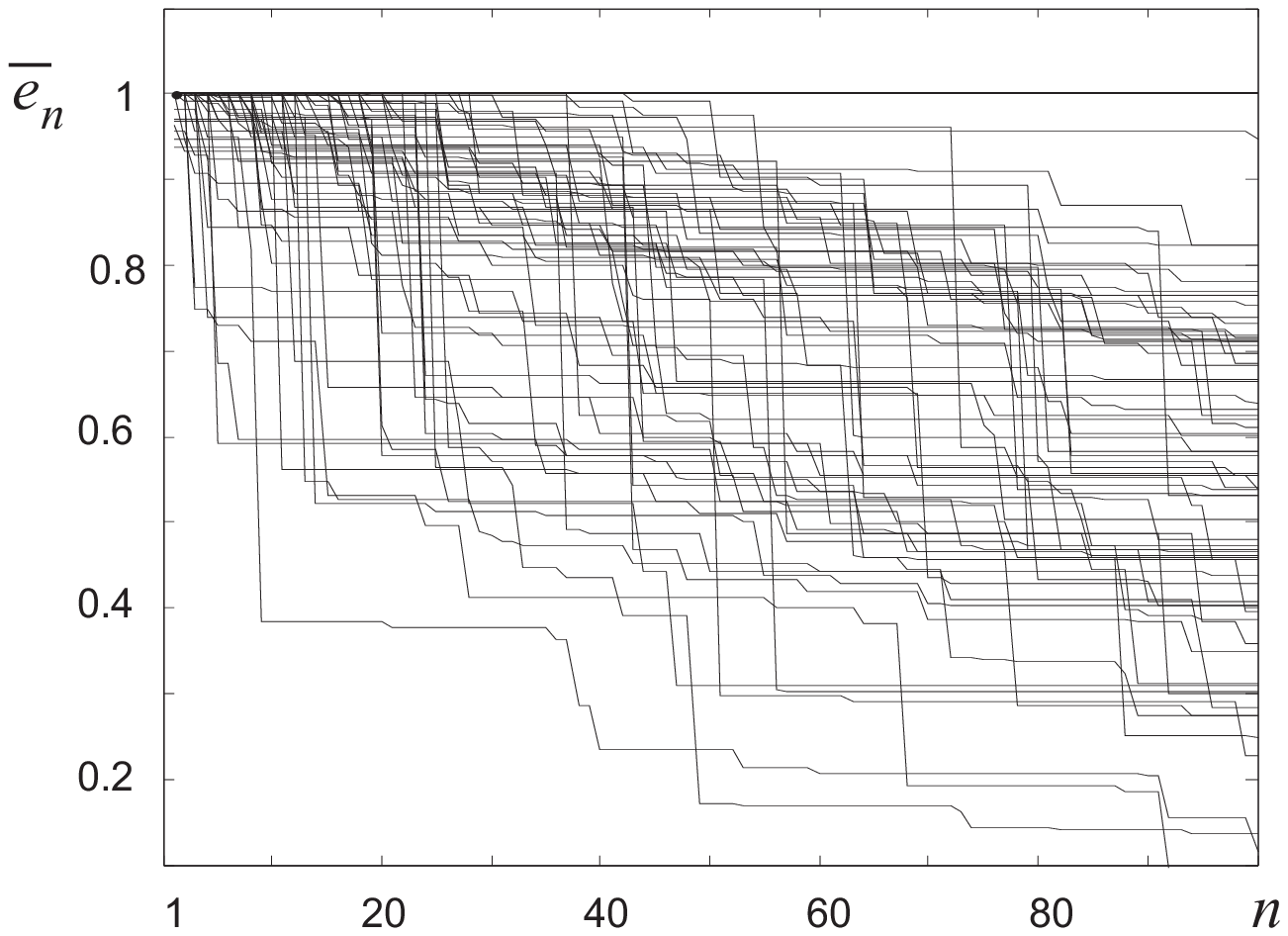}
\end{center}
\caption{Practical speed of convergence of function approximators
that use greedy algorithm (middle panel) and Monte-Carlo based
random choice of basis functions (bottom panel). The target function
is shown on the top panel. }\label{fig:example}
\end{figure}

In order to illustrate the main difference between greedy and RVFL
approximators, we consider the following example in which a simple
function is approximated by both methods, greedy approximation
(\ref{eq:approximation})--(\ref{eq:Jones_iteration}) and
approximation based on the Monte-Carlo integration
(\ref{eq:approximation:2})--(\ref{eq:rate:MC}). Let $f(x)$ be
defined as follows:
\[
f(x)=0.2 e^{-(10x-4)^2} + 0.5e^{-(80x-40)^2}+0.3e^{-(80x-20)^2}
\]
The function $f(x)$ is shown in Fig. \ref{fig:example}, top
panel. Clearly, $f(\cdot)$
belongs to the convex hull of $G$, and hence to its closure.

First, we implemented greedy approximation
(\ref{eq:approximation})--(\ref{eq:Jones_iteration}) in which we
searched for $g_n$ in the following set of functions
\[
G=\{e^{-(w^T x +
b)^2}\},
\]
where $w\in[0,200]$, $b\in[-100,0]$. The procedure for constructing $f_n$ was as follows.
Assuming $f_0(x)=0$, $e_0=-f$  we started with searching for $w_1$, $b_1$ such that
\begin{equation}\label{eq:exaple_greedy:1}
\begin{split}
&\langle0-f(x),g(w_1x+b)-f(x)\rangle=\\
&-\langle f(x),g(w_1x+b)\rangle+\|f(x)\|^2 < \varepsilon.
\end{split}
\end{equation}
where $\varepsilon$ was set to be small ($\varepsilon=10^{-6}$ in
our case). When searching for a solution of
(\ref{eq:exaple_greedy:1}) (which exists because the function $f$ is
in the convex hull of $G$ \cite{Jones:1992}), we did not utilize any
specific optimization routine. We sampled the space of parameters
$w_i$, $b_i$ randomly and picked the first values of $w_i$, $b_i$
which satisfy (\ref{eq:exaple_greedy:1}). Integral
(\ref{eq:exaple_greedy:1}) was evaluated in quadratures over a
uniform grid of $1000$ points in $[0,1]$.

The values of $\alpha_1$ and the function $f_1$ were chosen in
accordance with (\ref{eq:Jones_iteration}) with $M''=2$, $M'=1.5$
(these values are chosen to assure $M''>M'> \sup_{g} \|g\|+\|f\|$).
The iteration was repeated, resulting in the following sequence of
functions
\[
\begin{split}
f_n(x)&=\sum_{i=1}^n c_i g(w_i^{T}x+b_i), \\ c_i&=\alpha_i(1-\alpha_{i+1})(1-\alpha_{i+2})\cdots(1-\alpha_{n})
\end{split}
\]

Evolution of the normalized approximation error
\begin{equation}\label{eq:exaple_error_normalized}
\bar{e}_n=\frac{e_n^2}{\|f\|^2}=\frac{\|f_n-f\|^2}{\|f\|^2}
\end{equation}
for $100$ trials is shown in Fig. \ref{fig:example} (middle panel).
Each trial consisted of $100$ iterations
(\ref{eq:approximation})--(\ref{eq:Jones_iteration}), thus leading
to the networks of $100$ elements at the $100$th step. We observe
that the values of $\bar{e}_n$ monotonically decrease as $O(1/n)$,
with the behavior of this approximation procedure consistent across
trials.
% convergence to zero isn't shown

%{\bf Ivan, could you
%comment how you did greedy optimization and how many approximation
%functions you used ($n$ values)?  This will be very useful for not
%only for clarity of experiment description but also to confirm the
%fact that the greedy approximation is more efficient as it uses less
%approximating functions than the Monte-Carlo approximation for the
%same accuracy of approximation. As far as I understand, the
%Monte-Carlo approx. always uses $n$ approximating functions as the
%sample size, correct?}

Second, we implemented an approximator based on the Monte-Carlo
integration.  At the $n$th step of the approximation procedure we
pick randomly an element from $G$, where $w\in[0,200]$,
$b\in[-200,200]$ (uniform distribution). After an element is
selected, we add it to the current pool of basis functions
\[
P_{n-1}=\{g(w_1^Tx+b_1),\dots,g(w_{n-1}x+b_{n-1})\}.
\]
Then the weights $c_i$ in the superposition
\[
f_n=\sum_{i=1}^n c_i g(w_i^Tx+b_i)
\]
are optimized so that $\|f_n-f\|\rightarrow\min$. Evolution of the
normalized approximation error $\bar{e}_n$
(\ref{eq:exaple_error_normalized}) over $100$ trials is shown in
Fig. \ref{fig:example} (bottom panel). As can be observed from the
figure,  even though the values of $\bar{e}_n$ form a monotonically
decreasing sequence, they are far from $1/n$, at least for $1\leq
n\leq 100$. Behavior across trials is not consistent, at least for
the networks smaller than $100$ elements, as indicated by a
significant spread among the curves.
% should we expect them to be close to 1/n or to O(1/n)?

\begin{figure}[!t]
\begin{center}
\includegraphics[width=0.85\columnwidth]{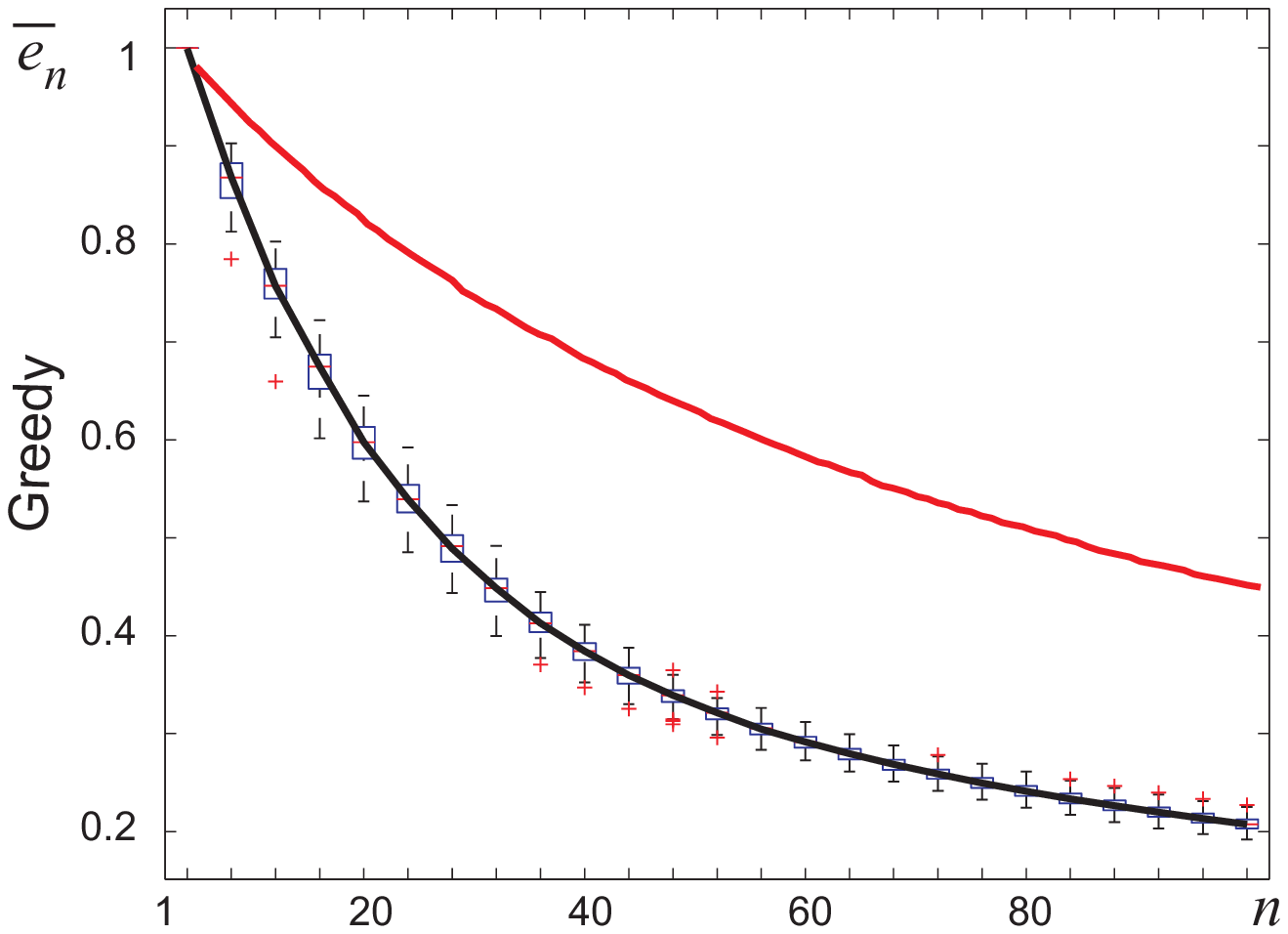}
\vskip 6mm
\includegraphics[width=0.85\columnwidth]{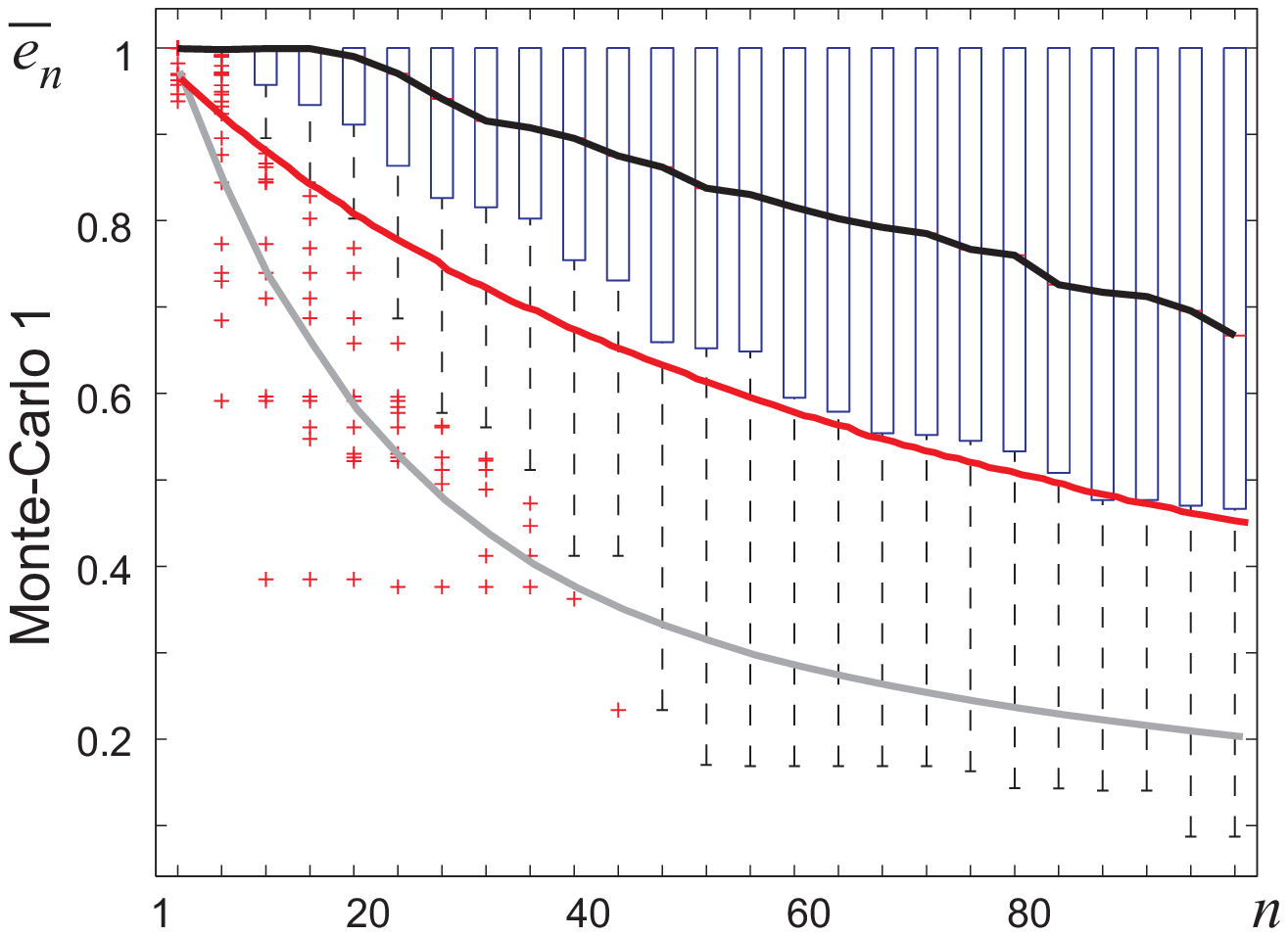}
\vskip 6mm
\includegraphics[width=0.85\columnwidth]{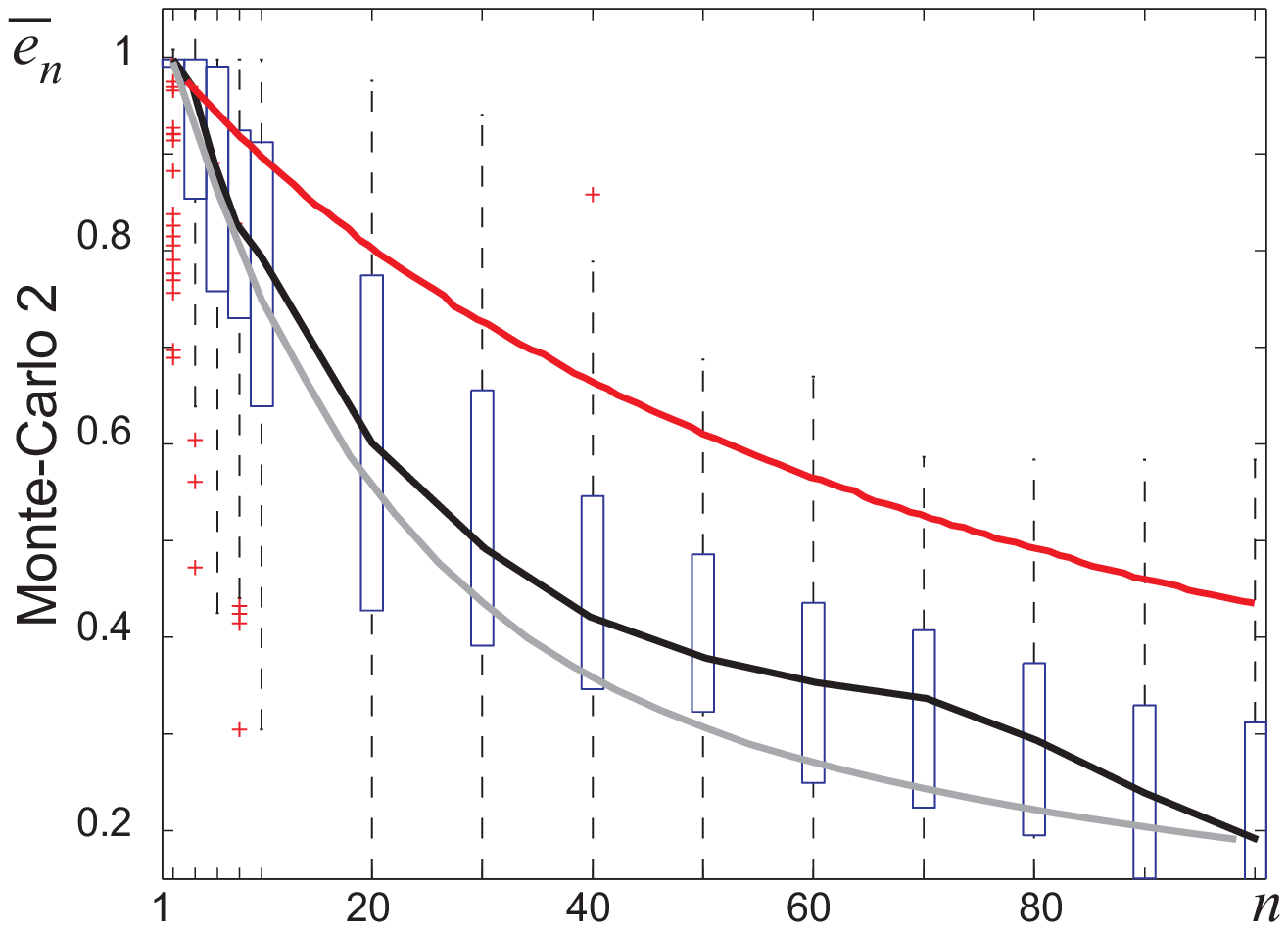}
\end{center}
\caption{Box plots of convergence rates for function approximators
that use greedy algorithm (top panel) and Monte-Carlo random choice
of basis functions (middle and bottom panels). The middle panel
corresponds to the case in which the basis functions leading to
ill-conditioning were discarded. The bottom panel shows performance
of the MLP trained by the method in \cite{Feldkamp98b} which is
effective at counteracting ill-conditioning while adjusting the
linear weights only. The red curve shows the upper bound for
$\bar{e}_n$ calculated in accordance with (\ref{eq:rate:greedy}). We
duplicated the average performance of the greedy algorithm (grey
solid curve) in the middle and bottom panels for convenience of
comparison.}\label{fig:example:2}
\end{figure}

Overall comparison of these two methods is provided in Fig.
\ref{fig:example:2}, in which the errors $\bar{e}_n$ are presented
in the form of a box plot. Black solid curves depict the median of
the error as a function of the number of elements, $n$, in the
network; blue boxes contain $50\%$ of the data points in all trials;
``whiskers'' delimit the areas containing $75\%$ of data, and red
crosses show the remaining part of the data. As we can see from
these plots, random basis function approximators, such as the RVFL
networks, mostly do not match performance of greedy approximators
for networks of reasonable size. Perhaps, employing integration
methods with variance minimization could improve the performance.
This, however, would amount to using prior knowledge about the
target function $f$, making it difficult to apply the RVFL networks
to problems in which the function $f$ is uncertain.

Now we demonstrate performance of an MLP trained to approximate this
target function. The NN is trained by a gradient based method
described in \cite{Feldkamp98b}. At first, the full network training
is carried out for several network sizes $n=20$, $40$, $60$, $80$
and $100$ and input samples randomly drawn from $x\in [0,1]$. The
values of $\bar{e}_n$ are $1.5\cdot10^{-4}$ for all the network
sizes (as confirmed in many training trials repeated to assess
sensitivity to weight initialization). This suggests that training
and performance of much smaller networks should be examined. The
networks with $n=2,4,6,8,10$ are trained, resulting in
$\bar{e}_n=0.5749,0.1416,0.0193,0.0011,0.0004$, respectively,
averaged over $100$ trials per the network size.  Next, we train
only the linear weights ($c_i$ in (\ref{eq:mlp})) of the MLP, fixing
the nonlinear weights $w_i$ and $b_i$ to random values. The results
for $\bar{e}_n$ averaged over $100$ trials are shown in Fig.
\ref{fig:example:2}, bottom panel (black curve). Remarkably, the
results of random basis network with $n=100$ are worse than those of
the MLP with $n\ge 4$ and full network training.  These results
indicate that both the greedy and the Monte-Carlo approximation
results shown in Fig. \ref{fig:example:2} are quite conservative.
Furthermore, the best of those two, i.e., the greedy
approximation's, can be dramatically improved by a practical
gradient based training.
%{\bf Ivan, I used RMSE*RMSE/(std(f)*std(f)) as $\bar{e}_n$. Further,
%you may want to show these $\bar{e}_n$ as stars or circles in Fig 1
%(middle panel). I didn't have time to run for other $n$, nor to
%repeat experiments 10 times for each $n$; I suspect that average
%$\bar{e}_n$ will behave better with respect to $n$. - This is fine
%at the moment and RMSE*RMSE/(std(f)*std(f)) is a discrete version of
%$\bar{e}_n$ indeed}

\addtolength{\textheight}{-1cm}   % This command serves to balance the column lengths

\section{Discussion}\label{sec:Discussion}

We just analyzed theoretically and illustrated on a simple example
what may happen if the basis for function approximation is chosen at
random.  We wish to discuss recent result presented in \cite{he05}
regarding the use of the random basis function approximators. We
choose this work because it is representative of a recent trend in
neural network control literature exemplified by
\cite{ok07}--\cite{Liuetal07}. In this trend, the purpose of one or
several neural networks implementing random basis is to account for
(and ideally - cancel asymptotically) an unknown bounded modeling
nonlinearity.

%  I COMMENTED THIS OUT BECAUSE IT JUST RESTATES WHAT IS SAID ABOVE: USE OF RANDOM BASIS. The approach described in \cite{he05}
%exploits the possibility to ensure arbitrarily good approximation of
%a nonlinear continuous function by NNs in which parameters of the
%hidden nodes are chosen at random, and only the weights of the
%linear layer are to be adjusted. This scheme guarantees {\it
%asymptotic} tracking of reference trajectories (up to some fixed
%error) provided that the state variables and outputs are bounded.

%The initial values of weights satisfy the boundedness
%assumption, and the goal of the adaptation is to keep everything
%bounded, i.e., uniform ultimate boundedness (UUB).
%In our opinion, the approach overly emphasizes the need to ensure
%boundedness of the estimates of the true values of the NN weights
%in the linear layer at the expense of performance and efficiency of the usage of approximating resources.

While ensuring that the tracking errors are bounded asymptotically,
the main theorem in \cite{he05} and its proof do not imply
performance improvement.
% THEY DON'T SPECIFY ANY PERFORMANCE IMPROVEMENT EXPLICITLY.
Instead the proof attempts to relate design parameters $\gamma_i$
with magnitudes of disturbances and weights of neural networks.
% WEIGHTS INSTEAD OF IDEAL WEIGHTS BECAUSE THEY USE WEIGHT DIFFERENCE IN LYAPUNOV FUNCTION, SO
% WEIGHTS SHOULD CONVERGE TO IDEAL WEIGHTS (IDEAL WEIGHTS IS A BAD CONCEPT BUT THAT'S FOR OUR ANOTHER PAPER :-)
Though the disturbance magnitude may indeed be known a priori, one
can not assume sufficiently small bounds on the values of weights
because the weights may need to be large in order to compensate for
residual modeling errors from randomly assigned basis functions.
Furthermore, the larger the weights or the farther the system of
basis functions from an orthogonal one (ill-conditioning), the more
time is needed for an adaptive system to converge into the desired
domain; see, e.g., \cite{French2000}. In fact, in the example of
Section \ref{sec:Example} we observed values of the hidden layer
weights as large as $200$. However, large bounds on the weights
force the control system designer to decrease design parameters
$\gamma_i$ which, in turn, results in an increase of the region of
uniform ultimate boundedness (UUB) (determined by equations
(A.5-A.9) in \cite{he05}). Ironically, the region of UUB in
\cite{he05} may not shrink to zero even in the ideal case of zero
disturbances.
% IVAN, I NOTICED THAT THERE'S NO WAY TO KILL DM BY INCREASED GAMMA6 IN (A.7)!
The UUB depending on such uncontrollable quantities as weights makes
it impossible to provide practically valuable guarantees of the
closed-loop system performance.
%WE DIDN'T TRAIN APPROXIMATORS IN AN ADAPTIVE SETTING, JUST IN MAPPING APPROXIMATION SETTING
%These both factors negatively
%affecting performance of an adaptive system were persistently
%observed in our benchmark example for the RVFL.

%%%%%%%%%%%%%%%%%%%%%%%%%%%%%%%%%%%%%%%%%%%%%%%%%%%%%%%%%%%%%%%%%%%%%%%%%%%%%%%%

%%%%%%%%%%%%%%%%%%%%%%%%%%%%%%%%%%%%%%%%%%%%%%%%%%%%%%%%%%%%%%%%%%%%%%%%%%%%%%%%
\section{Conclusion}\label{sec:Conclusion}

In this work we demonstrate that, despite increasing popularity of
random basis function networks in control literature, especially in
the domain of intelligent/adaptive control, one needs to pay special
attention to practical aspects that may affect performance of these
systems in applications.

First, as we analyzed in Section II and showed in our example,
although the rate of convergence of the random basis function
approximator is qualitatively similar to that of the greedy
approximator, the rate of the random basis function approximator is
achievable only when the number of elements in the network is
sufficiently large. Second, approximators which are motivated by the
Monte-Carlo integration method offer only {\it statistical} measure
of approximation quality.
% how about local minima which may also affect greedy approximators,
% hence their errors also become statistical?
In other words, small approximation errors are guaranteed here in
{\it probability}. This means that, for practical adaptive control
in which the RVFL networks are to model or compensate system
uncertainties, employment of a re-initialization with a supervisory
mechanism monitoring quality of the RVFL network is necessary.
Unlike network training methods that adjust both linear and
nonlinear weights of the network, such mechanism may have to be made
robust against numerical problems (ill-conditioning) which often
occurs in the Monte-Carlo method.

Our conclusion about the random basis function approximators is also
consistent with the following intuition. If the approximating
elements (network nodes) are chosen at random and not subsequently
trained, they are usually not placed in accordance with the density
of the input data. Though computationally easier than for nonlinear
parameters, training of linear parameters becomes ineffective at
reducing errors ``inherited" from the nonlinear part of the
approximator. Thus, in order to improve effectiveness of the random
basis function approximators one could combine unsupervised
placement of network nodes according to the input data density with
subsequent supervised or reinforcement learning values of the linear
parameters of the approximator.  However, such a combination of
methods is not-trivial because in adaptive control and modeling one
often has to be able to allocate approximation resources adaptively
-- and the full network training seems to be the natural way to
handle such adaptation.

% BASED ON MY EXPERIENCE, I AM CONFIDENT THAT FULL NET TRAINING WILL DELIVER GOOD RESULTS, SO WE DON'T NEED TO
% SOUND LIKE WE HAVEN'T TRIED MUCH BIGGER PROBLEMS; THE STRENGTH OF THE PAPER IS IN THEORETICAL ANALYSIS, NOT
% EXPERIMENTS.
%Our results also show that performance of classical backpropagation networks
%in the considered low-dimensional benchmark approximation task is on par with
%the greedy approximation, and they both outperform the RVFL networks,
%at least within the class of networks in which the number of elements does not exceed $100$.
%Whether such trend remains in higher dimensions and for a wider range of functions is a
%subject of study to be presented in the full paper.

%{\bf Ivan, could one say that greedy approximation also offers
%approximation quality only statistically because of using finite
%data set for training (sampling of data) -- and in that sense no
%different from the Monte-Carlo or backpropagation?  Just wonder if
%such a confusion is possible.}

%%%%%%%%%%%%%%%%%%%%%%%%%%%%%%%%%%%%%%%%%%%%%%%%%%%%%%%%%%%%%%%%%%%%%%%%%%%%%%%%

\section*{Acknowledgment}

The authors are grateful to Prof. A.N. Gorban for  useful comments and numerous technical discussions during preparation of this work. The first author's research was
supported by a Royal Society International Joint Project grant, and
partially supported by RFBR grant 8-08-00103-a.

\bibliographystyle{plain}
\bibliography{isic09critica3}

\end{document}